\documentclass[11pt,a4paper]{article}
\usepackage[hyperref]{emnlp2020}

\usepackage{times}
\usepackage{latexsym}

\usepackage{microtype}

\usepackage{times}
\usepackage{latexsym}
\usepackage{xspace}
\usepackage{enumitem}

\usepackage{url}

\usepackage{booktabs} 
\usepackage{diagbox}
\usepackage{multirow}
\usepackage{balance}
\usepackage{amsfonts}
\usepackage{amsmath}
\usepackage{amssymb}
\usepackage{graphicx}
\usepackage{subfigure}
\usepackage{microtype}
\usepackage{xspace}
\usepackage{wrapfig,lipsum}
\usepackage{makecell}
\usepackage[T1]{fontenc}
\usepackage{soul}
\usepackage{dsfont}
\usepackage{tabularx, booktabs}

\aclfinalcopy 

\newcommand\DNAME{ADMIN}

\newcommand{\eat}[1]{\ignorespaces}

\title{Very Deep Transformers for Neural Machine Translation}

\author{
Xiaodong Liu$^\dagger$, Kevin Duh$^\ddagger$, Liyuan Liu$^\S$ and Jianfeng Gao$^\dagger$
 \\ 
  $^\dagger$ Microsoft Research ~~~~
  $^\ddagger$ Johns Hopkins University \\
  $^\S$ University of Illinois at Urbana-Champaign\\
  {\tt \{xiaodl,jfgao\}@microsoft.com} \\
  \tt kevinduh@cs.jhu.edu, ll2@illinois.edu
}

\date{}

\begin{document}
\maketitle

\begin{abstract}

We explore the application of very deep Transformer models for Neural Machine Translation (NMT). 
Using a simple yet effective initialization technique that stabilizes training, we show that it is feasible to build standard Transformer-based models with up to 60 encoder layers and 12 decoder layers. 
These deep models outperform their baseline 6-layer counterparts by as much as 2.5 BLEU, and achieve
new state-of-the-art benchmark results on 
WMT14 English-French (43.8 BLEU and 46.4 BLEU with back-translation) and WMT14 English-German (30.1 BLEU).
To facilitate further research in Very Deep Transformers for NMT, we release the code and models: \url{https://github.com/namisan/exdeep-nmt}.
\end{abstract}


\section{Introduction}
\label{sec:intro}

The capacity of a neural network influences its ability to model complex functions. 
In particular, it has been argued that \textit{deeper} models are conducive to more expressive features \cite{bengio09book}.
Very deep neural network models have proved successful in computer vision \cite{he2016resnet,srivastava2015deepnn} and text classification \cite{conneau2017deepclassification,minaee2020deep}.
In neural machine translation (NMT), however, current state-of-the-art models such as the Transformer typically employ only 6-12 layers \cite{wmt19edinburg, wmt19ms, wmt19fair}.

Previous work has shown that it is difficult to train deep Transformers, such as those over 12 layers \cite{bapna2018training}.
This is due to optimization challenges: the variance of the output at each layer compounds as they get deeper, leading to unstable gradients and ultimately diverged training runs.

In this empirical study, we re-investigate whether deeper Transformer models are useful for NMT. 
We apply a recent initialization technique called ADMIN \cite{liu2020deeptransformer}, which remedies the variance problem. This enables us train Transformers that are significantly deeper, e.g. with 60 encoder layers and 12 decoder layers.\footnote{We choose to focus on this layer size since it results in the maximum model size that can fit within a single GPU system. The purpose of this study is to show that it is feasible for most researchers to experiment with very deep models; access to massive GPU budgets is not a requirement.}

In contrast to previous research, we show that it is indeed feasible to train the standard\footnote{Note there are architectural variants that enable deeper models \cite{wang2019preln,Nguyen2019fixnorm}, discussed in Sec \ref{sec:prelimiary}. We focus on the standard architecture here.} Transformer \cite{vaswani2017attention} with many layers. 
These deep models significantly outperform their 6-layer baseline, with up to 2.5 BLEU improvement. Further, they obtain state-of-the-art on the WMT'14 EN-FR and WMT'14 EN-DE benchmarks.

\begin{figure}[t]
    \centering
    \includegraphics[width=0.75\linewidth]{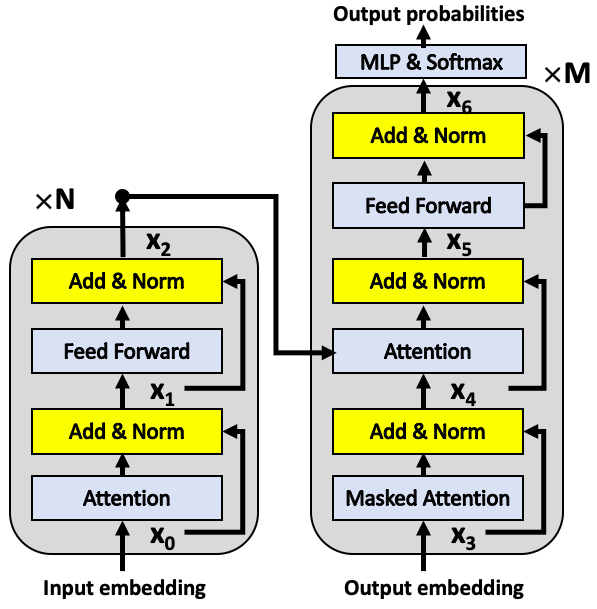}
    \caption{Transformer model}
    \label{fig:transformer}
    \vspace{-4mm}
\end{figure}

\section{Background}
\label{sec:prelimiary}

We focus on the Transformer model \cite{vaswani2017attention}, shown in Figure \ref{fig:transformer}. 
The encoder consists of $N$ layers/blocks of attention + feed-forward components. The decoder consists of $M$ layers/blocks of masked-attention, attention, and feed-forward components. 
To illustrate, the input tensor $\mathbf{x_{i-1}}$ at the encoder is first transformed by a multi-head attention mechanism to generate the tensor $f_{ATT}(\mathbf{x_{i-1}})$. This result is added back with $\mathbf{x_{i-1}}$ as a residual connection, then layer-normalization ($f_{LN}(\cdot)$) is applied to generate the output:
\begin{math}
    \mathbf{x_{i}}=f_{LN}(\mathbf{x_{i-1}} + f_{ATT}(\mathbf{x_{i-1}}))
\end{math}.
Continuing onto the next component, $\mathbf{x_{i}}$ is passed through a feed-forward network $f_{FF}(\cdot)$, and is again added and layer-normalized to generate the output tensor: 
$\mathbf{x_{i+1}}=f_{LN}(\mathbf{x_{i}} + f_{FF}(\mathbf{x_{i}}))$.
Abstractly, the output tensor at each Add+Norm component in the Transformer (Figure \ref{fig:transformer}) can be expressed as: 
\begin{equation}
\label{eq:postnorm}
    \mathbf{x_{i}}=f_{LN}(\mathbf{x_{i-1}} + f_{i}(\mathbf{x_{i-1}}))
\end{equation}
where $f_{i}$ represents a attention, masked-attention, or feed-forward subnetwork. 
This process repeats $2 \times N$ times for a $N$-layer encoder and $3 \times M$ times for a $M$-layer decoder.
The final output of the decoder is passed through a softmax layer which predicts the probabilities of output words, and the entire network is optimized via back-propagation. 

Optimization difficulty has been attributed to vanishing gradient, despite layer normalization \cite{xu2019improveln} providing some mitigation. 
The lack of gradient flow between the decoder and the {\em lower} layers of the encoder is especially problematic; this can be addressed with short-cut connections \cite{bapna2018training,he2018layer-wise}.
An orthogonal solution is to swap the positions of layerwise normalization $f_{LN}$ and subnetworks $f_{i}$ within each block \cite{Nguyen2019fixnorm,domhan2018ananmt,chen2018-best-adv-nmt} by:  
\begin{math}
\label{eq:prenorm}
    \mathbf{x_{i}}=f_{i}(\mathbf{x_{i-1}} + f_{LN}(\mathbf{x_{i-1}}))
\end{math}
This is known as pre-LN (contrasted with post-LN in Eq.~\ref{eq:postnorm}), and has been effective in training networks up to 30 layers \cite{wang2019preln}.\footnote{The 96-layer GPT-3 \cite{brown2020gpt3} uses pre-LN.}

However, it has been shown that post-LN, if trained well, can outperform pre-LN \cite{liu2020deeptransformer}. 
Ideally, we hope to train a standard Transformer without additional architecture modifications. 
In this sense, our motivation is similar to that of \citet{wu2019depth}, which grows the depth of a standard Transformer in a stage-wise fashion.

\section{Initialization Technique}
\label{sec:dnmt}

The initialization technique ADMIN \cite{liu2020deeptransformer} we will apply here reformulates Eq.~\ref{eq:postnorm} as:
\vspace{-2mm}
\begin{equation}
    \mathbf{x_{i}}=f_{LN}(\mathbf{x_{i-1}}\cdot \mathbf{\omega_i} + f_{i}(\mathbf{x_{i-1}}))
    \label{eq:admin}
\end{equation}
where $\mathbf{\omega_i}$ is a constant vector that is element-wise multiplied to $\mathbf{x_{i-1}}$ in order to balance the contribution against $f_{i}(\mathbf{x_{i-1}})$. 
The observation is that in addition to vanishing gradients, the unequal magnitudes in the two terms 
$\mathbf{x_{i-1}}$ and $f_{i}(\mathbf{x_{i-1})}$ is the main cause of instability in training.
Refer to \cite{liu2020deeptransformer} for theoretical details.\footnote{Note that paper presents results of 18-layer Transformers on the WMT'14 En-De, which we also use here. Our contribution is a more comprehensive evaluation.} 

ADMIN initialization involves two phases:
At the {\bf Profiling} phase, we randomly initialize the model parameters using default initialization, set $\mathbf{\omega_i}=1$, and perform one step forward pass in order to compute the output variance of the residual branch $Var[f(\mathbf{x_{i-1})}]$ at each layer.\footnote{We estimate the variance with one batch of 8k tokens.} In the {\bf Training} phase, we fix $\mathbf{\omega_i}~=~\sqrt{\sum_{j < i}{Var[f(\mathbf{x_{j-1}})]}}$, and then train the model using standard back-propagation. 
After training finishes, $\mathbf{\omega_i}$ can be folded back into the model parameters to recover the standard Transformer architecture. 
This simple initialization method is effective in ensuring that training does not diverge, even in deep networks. 
\section{Experiments}
\label{sec:exp}

\begin{table*}[h]
    \centering
    \begin{tabular}{|l||c|c|c|c|c|c|c|c|c|c|}
        \toprule
        & \multicolumn{5}{c|}{WMT'14 English-French (FR)} & \multicolumn{5}{c|}{WMT'14 English-German (DE)} \\
        {\bf Model} & {\bf \#param} &{\bf T}$\downarrow$ &{\bf M}$\uparrow$ &{\bf BLEU}$\uparrow$ & {$\Delta$}& {\bf \#param} &{\bf T}$\downarrow$ &{\bf M}$\uparrow$ & {\bf BLEU}$\uparrow$ & {$\Delta$} \\
        \midrule
        
        6L-6L Default & 67M & 42.2 & 60.5 & 41.3 & - & 61M & 54.4 & 46.6 &  27.6 & - \\ \hline
        6L-6L \DNAME & 67M & 41.8 & 60.7 & 41.5 & 0.2 & 61M & 54.1 & 46.7 &  27.7 & 0.1 \\ \hline

        \hline
        60L-12L Default & 262M & \multicolumn{4}{c|}{diverge} & 256M & \multicolumn{4}{c|}{diverge}  \\ \hline
        60L-12L \DNAME & 262M & {\bf 40.3} & {\bf 62.4} & {\bf 43.8} & 2.5 & 256M & {\bf 51.8} & {\bf 48.3} &  {\bf 30.1} & 2.5 \\ \hline

    \end{tabular}
    \caption{Test results on WMT'14 benchmarks, in terms of TER ({\bf T}$\downarrow$), METEOR ({\bf M}$\uparrow$), and BLEU. {$\Delta$} shows difference in BLEU score against baseline 6L-6L Default. Best results are boldfaced. 60L-12L ADMIN outperforms 6L-6L in all metrics with statistical significance ($p<0.05$). Following convention, BLEU is computed by \texttt{multi-bleu.perl} via the standardized tokenization of the publicly-accessible dataset. }
    \label{tab:main_result}
    \vspace{-4mm}
\end{table*}

Experiments are conducted on standard WMT'14 English-French (FR) and English-German (DE) benchmarks. 
For FR, we mimic the setup\footnote{\url{https://github.com/pytorch/fairseq/blob/master/examples/translation/prepare-WMT'14en2fr.sh}} of \cite{ott-etal-2018-scaling}, with 36M training sentences and 40k subword vocabulary.
We use the provided 'valid' file for development and newstest14 for test.
For DE, we mimic the setup\footnote{\url{https://github.com/tensorflow/tensor2tensor/blob/master/tensor2tensor/data_generators/translate_ende.py}} of \cite{so19evolved}, with 4.5M training sentences, 32K subword vocabulary, newstest2013 for dev, and newstest2014 for test.

We adopt the hyper-parameters of the Transformer-based model \cite{vaswani2017attention} as implemented in FAIRSEQ \cite{ott2019fairseq}, i.e. 512-dim word embedding, 2048 feed-forward model size, and 8 heads, but vary the number of layers. RAdam \cite{liu2019radam} is our optimizer.\footnote{For FR, \#warmup steps is 8000,  max \#epochs is 50, and peak learning rate is 0.0007. For DE, \#warmup steps is 4000,  max \#epochs is 50, and learning rate is 0.001. Max \#tokens in each batch is set to 3584 following \cite{ott2019fairseq}.} 

\begin{table}[ht]
    \centering
    \begin{tabular}{|l|c|c|}
    \hline
        BLEU via \texttt{multi-bleu.perl} & FR & DE \\\hline
        60L-12L ADMIN & {\bf 43.8}  & {\bf 30.1}\\
        \cite{wu2019depth}  & 43.3 & 29.9 \\
        \cite{wang2019preln} & - & 29.6\\
        \cite{wu19dycnn}  & 43.2 & 29.7 \\
        \cite{ott-etal-2018-scaling} & 43.2 & 29.3 \\
        \cite{vaswani2017attention} & 41.8 & 28.4 \\
        \cite{so19evolved}   & 41.3 & 29.8 \\    
        \cite{gehring2017}  & 40.5 & 25.2\\
        \hline\hline
        BLEU via \texttt{sacreBLEU.py} & FR & DE \\\hline
        60L-12L ADMIN & {\bf 41.8}  & {\bf 29.5} \\        
        \cite{ott-etal-2018-scaling} & 41.4 & 28.6 \\
        \cite{so19evolved}   & n/a & 29.2 \\    
        \hline
    \end{tabular}
    \caption{State-of-the-Art on WMT'14 EN-FR/EN-DE}
    \label{tab:sota}
\vspace{-4mm}    
\end{table}



\begin{figure}[th]
\centering  
\subfigure[Train set perplexity: Default vs ADMIN]{
\includegraphics[width=0.9\linewidth]{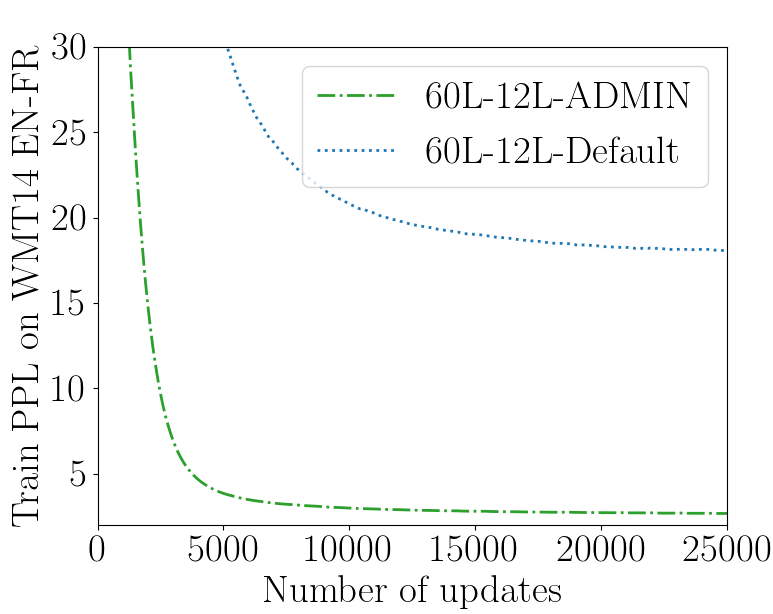}
}
\subfigure[Dev set perplexity: different ADMIN models]{\includegraphics[width=0.9\linewidth]{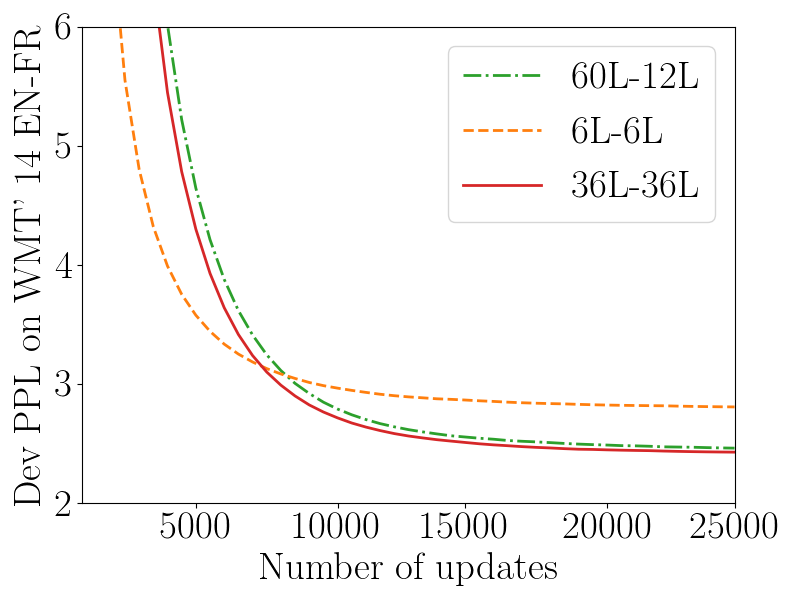}}
\caption{Learning curve}
	\label{fig:learning}	
\end{figure}

\paragraph{Main Result:}
Our goal is to explore whether very deep Transformers are feasible and effective. 
We compare: (a) \textbf{6L-6L}: a baseline Transformer Base with 6 layer encoder and 6 layer decoder, vs. (b) \textbf{60L-12L}: A deep transformer with 60 encoder layers and 12 decoder layers.\footnote{We use ``(N)L-(M)L" to denote that a model has N encoder layers and M decoder layers. N \& M are chosen based on GPU (16G) memory constraint. For reproducibility and simplicity, we focused on models that fit easily on a single GPU system. Taking FR as an example, it takes 2.5 days to train 60L-12L using one DGX-2 (16 V100's), 2 days to train a 6L-6L using 4 V100's.}
For each architecture, we train with either default initialization \cite{glorot2010init} or ADMIN initialization.

The results in terms of BLEU \cite{bleu}, TER \cite{snover06}, and METEOR \cite{lavie07meteor} are reported in Table \ref{tab:main_result}. 
Similar to previous work \cite{bapna2018training}, we observe that deep 60L-12L Default diverges during training. 
But the same deep model with ADMIN successfully trains and impressively achieves 2.5 BLEU improvement over the baseline 6L-6L Default in both datasets. 
The improvements are also seen in terms of other metrics: in EN-FR, 60L-12L ADMIN outperforms the 6L-6L models in TER (40.3 vs 42.2) and in METEOR (62.4 vs 60.5). All results are statistically significant ($p<0.05$) with a 1000-sample bootstrap test \cite{clark2011boost-smt}.

These results indicate that it is feasible to train standard (post-LN) Transformers that are very deep.\footnote{Note: the pre-LN version does train successively on 60L-12L and achieves 29.3 BLEU in DE \& 43.2 in FR. It is better than 6L-6L but worse than 60L-12L ADMIN.}
These models achieve state-of-the-art results in both datasets. The top results in the literature are compared in Table~\ref{tab:sota}.\footnote{The table does not include systems that use extra data.} We list BLEU scores computed with \texttt{multi-bleu.perl} on the tokenization of the downloaded data (commonly done in previous work), and with \texttt{sacrebleu.py} (version: tok.13a+version.1.2.10). which allows for a safer token-agnostic evaluation \cite{post2018sacrebleu}.

\paragraph{Learning Curve:}

We would like to understand why 60L-12L ADMIN is doing better from the optimization perspective. 
Figure~\ref{fig:learning} (a) plots the learning curve comparing ADMIN to Default initialization. We see that Default has difficulty decreasing the training perplexity; its gradients hit NaN, and the resulting model is not better than a random model.
In Figure~\ref{fig:learning} (b), we see that larger models (60L-12L, 36L-36L) are able obtain lower dev perplexities than 6L-6L, implying that the increased capacity does lead to better generalization.

\begin{figure}[t]
\centering  
\subfigure[Word accuracy according to frequency in the training data]{
\includegraphics[width=0.99\linewidth]{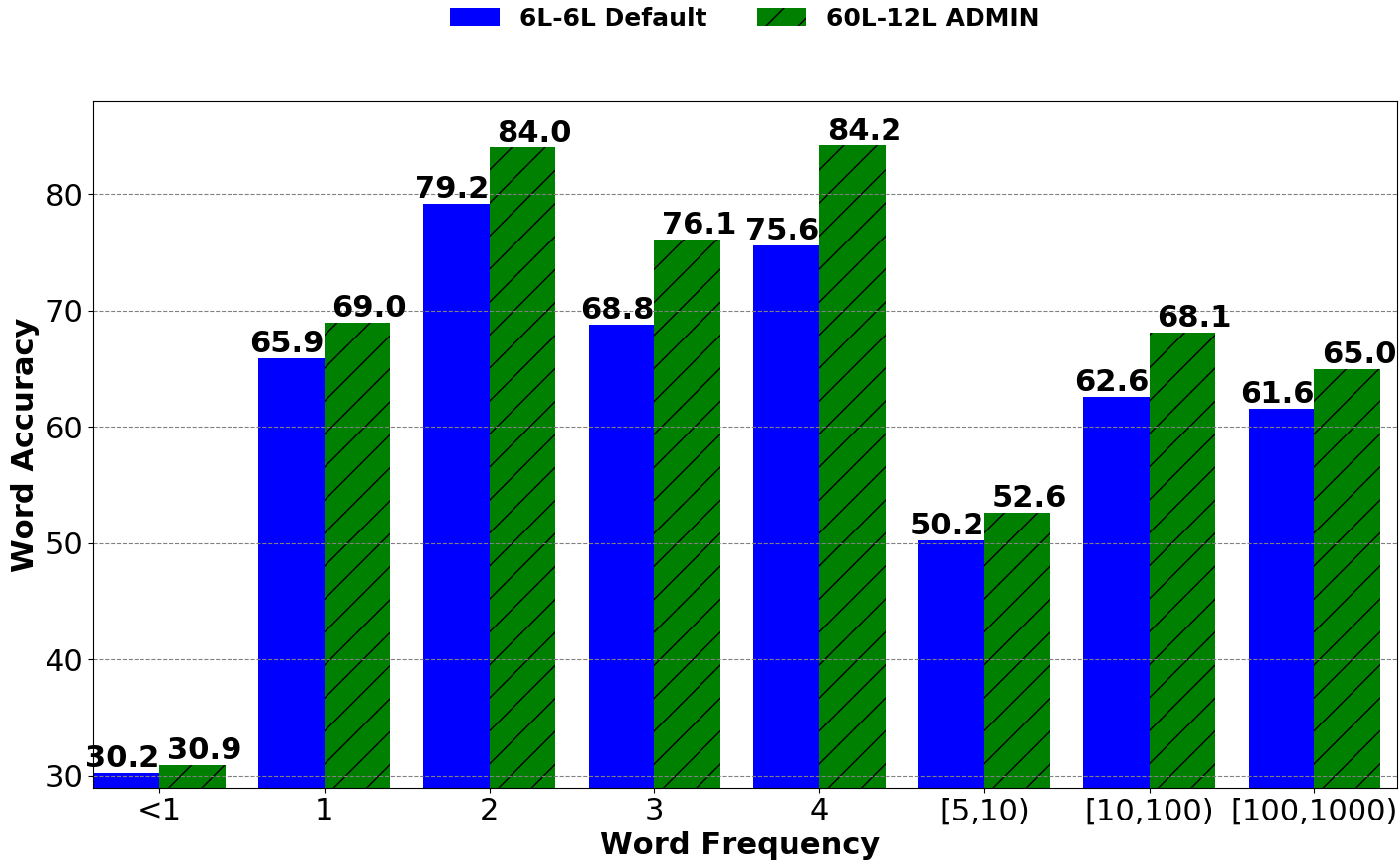}
}
\subfigure[BLEU scores according to sentence length.]{\includegraphics[width=0.98\linewidth]{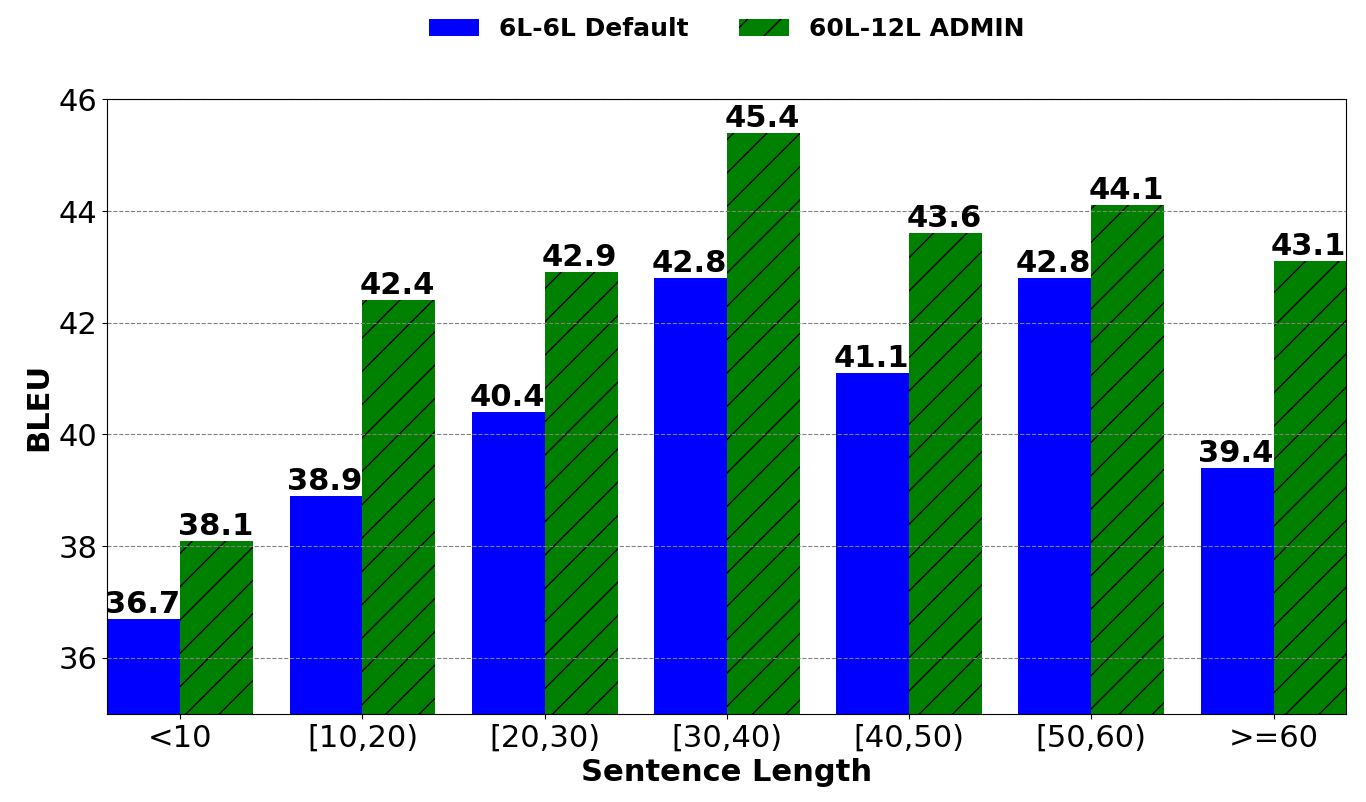}}
\caption{Fine-grained Error Analysis: note the deep model performs better across the board, indicating that it helps translation in general.}
	\label{fig:error_analysis}	
\end{figure}

\paragraph{Fine-grained error analysis:}
We are also interested in understanding how BLEU improvements are reflected in terms of more nuanced measures. For example, do the deeper models particularly improve translation of low frequency words?  Do they work better for long sentences? The answer is that the deeper models appear to provide improvements generally across the board (Figure~\ref{fig:error_analysis}).\footnote{Computed by \texttt{compare-mt} \cite{neubig2019-compare-nmt}.}

%
%

\begin{table}[t]
    \centering
    \begin{tabular}{|lc|c|c|c|c|c|c|c|}
    \hline
    {\bf Model} & {\small BLEU} & a & b & c & d & e & f & g  \\
    \hline\hline
    a:6L-6L  & 41.5  &   & - & - & - & - & - & - \\\hline
    b:12L-12L & 42.6 & + &   & - & - & - & - & - \\\hline
    c:24L-12L &  43.3 & + & + &   & = & - & = & = \\\hline
  d:48L-12L &  43.6  & + & + & = &   & = & = & + \\\hline
  e:60L-12L &  43.8  & + & + & + & = &   & = & + \\\hline
  f:36L-36L & 43.7   & + & + & = & = & = &   & + \\\hline
  g:12L-60L & 43.1  & +  & + & = & - & - & - & \\\hline
    \end{tabular}
    \caption{BLEU comparison of different encoder and decoder layers (using ADMIN initialization, on WMT'14 EN-FR). In the matrix, each element (i,j) indicates if the model in row~i significantly outperforms the model in column~j (+), under-performs j (-), or has no statistically significant difference (=). }
    \label{tab:layer_ablation}
\vspace{-4mm}    
\end{table}

\paragraph{Ablation Studies:}
We experimented with different number of encoder and decoder layers, given the constraint of a 16GB GPU.
Table~\ref{tab:layer_ablation} shows the pairwise comparison of models. 
We observe that 60L-12L, 48L-12L, and 36L-36L are statistically tied for best BLEU performance. 
It appears that deeper encoders are more worthwhile than deeper decoders, when comparing 60L-12L to 12L-60L, despite the latter having more parameters.\footnote{Recall from Figure \ref{fig:transformer} that each encoder layer has 2 subnetwork components and each decoder layer has 3 components.}

We also experiment with wider networks, starting with a 6L-6L Transformer-Big (1024-dim word embedding, 4096 feed-forward size, 16 heads) and doubling its layers to 12L-12L. The BLEU score on EN-FR improved from 43.2 to 43.6 (statistically significant, $p<0.05$).
A 24L-12L Transformer with BERT-Base like settings (768-dim word embedding, 3072 feed-forward size, 12 heads) obtain \textbf{44.0} BLEU score on WMT'14 EN-FR.
This shows that increased depth also helps models that are already relatively wide.

\paragraph{Back-translation}
We investigate whether deeper models also benefit when trained on the large but potentially noisy data such as back-translation. We follow the back-translation settings of \cite{edunov2018back-nmt} and generated additional 21.8M translation pairs for EN-FR. The hyperparameters are the same as the one without back-translation as introduced in \cite{edunov2018back-nmt}, except for an up-sampling rate 1 for EN-FR.

Table~\ref{tab:admin-bt} compares the ADMIN 60L-12L and ADMIN 36L-12L-768D model \footnote{It is BERT-base setting with 768-dim word embedding, 3072 feed-froward size and 12 heads.} with the default big transformer architecture (6L-6L) which obtains states-of-the-art results \cite{edunov2018back-nmt}. We see that with back-translation, both ADMIN 60L-12L + BT and ADMIN 36L-12L-768D still significantly outperforms its baseline ADMIN 60L-12L. Furthermore, ADMIN 36L-12L-768D achieves new state-of-the-art benchmark results on WMT'14 English-French (46.4 BLEU and 44.4 sacreBLEU \footnote{BLEU+case.mixed+lang.en-fr+numrefs.1+smooth.exp+test.wmt14+tok.13a+version.1.2.10}).

\begin{table}[ht]
    \centering
    \begin{tabular}{|l|c|}
    \hline
        BLEU via \texttt{multi-bleu.perl} & FR  \\\hline
        36L-12L-768D ADMIN + BT & {\bf 46.4}  \\
        60L-12L ADMIN + BT & 46.0  \\
       BT \cite{edunov2018back-nmt} & 45.6 \\
        60L-12L ADMIN & 43.8  \\ 
        \hline\hline
        BLEU via \texttt{sacreBLEU.py} & FR \\\hline
        36L-12L-768D ADMIN + BT & {\bf 44.4}  \\
        60L-12L ADMIN + BT & 44.1  \\
        60L-12L ADMIN & 41.8  \\
        BT \cite{edunov2018back-nmt} & - \\
        \hline
    \end{tabular}
    \caption{Back-translation results on WMT'14 EN-FR.}
    \label{tab:admin-bt}
\vspace{-4mm}    
\end{table}

\section{Conclusion}
\label{sec:conclusion}
We show that it is feasible to train Transformers at a depth that was previously believed to be difficult. 
Using ADMIN initialization, we build Transformer-based models of 60 encoder layers and 12 decoder layers. On WMT'14 EN-FR and WMT'14 EN-EN, these deep models outperform the conventional 6-layer Transformers by up to 2.5 BLEU, and obtain state-of-the-art results.

We believe that the ability to train very deep models may open up new avenues of research in NMT, including:
(a) Training on extremely large but noisy data, e.g. back-translation \cite{edunov2018back-nmt} and adversarial training \cite{cheng2019robust,liu2020alum}, to see if it can be exploited by the larger model capacity. 
(b) Analyzing the internal representations, to see if deeper networks can indeed extract higher-level features in syntax and semantics \cite{belinkov19}. (c) Compressing the very deep model via e.g. knowledge distillation \cite{kim-rush-2016-sequence}, to study the trade-offs between size and translation quality. (d) Analyzing how deep models work \cite{allen2020backward} in theory.
\section*{Acknowledgments}
We thank Hao Cheng, Akiko Eriguchi, Hany Hassan Awadalla and Zeyuan Allen-Zhu for valuable discussions. 

\bibliography{ref}
\bibliographystyle{acl_natbib}
\end{document}